\documentclass[letterpaper, 10 pt, conference]{ieeeconf}  % Comment this line out if you need a4paper

\IEEEoverridecommandlockouts                              % This command is only needed if 
\usepackage{graphics} % for pdf, bitmapped graphics files
\usepackage{amsmath} % assumes amsmath package installed
\usepackage{amssymb}  % assumes amsmath package installed
\usepackage{etoolbox}
\usepackage[ruled]{algorithm2e} % For algorithms
\usepackage{wrapfig}
\usepackage{lipsum}

\usepackage{textcase}
\usepackage[tablename=TABLE,font={footnotesize}]{caption}
\DeclareCaptionTextFormat{up}{\MakeTextUppercase{#1}}
\SetKwProg{Fn}{Function}{}{}
\SetKwComment{Comment}{$\triangleright$\ }{}

\graphicspath{{resources/}}

\usepackage[table]{xcolor}
\usepackage{graphicx}
\usepackage{svg}
\usepackage[normalem]{ulem}
\usepackage{siunitx}
\DeclareSIUnit{\rad}{rad}
\usepackage{cite}
\let\labelindent\relax
\usepackage{enumitem}
\usepackage{comment}
\usepackage{booktabs}
\usepackage{multirow}

\newcommand{\bh}[1] {{\color{blue} #1} }

\newcommand{\mtf}{\textsc{Baseline}\xspace}
\newcommand{\mts}{\textsc{RL}\xspace}
\newcommand{\earl}{\textsc{EARL}\xspace}

\usepackage{hyperref}

\def\b#1{\textcolor{blue}{#1}}

\title{\bf%\LARGE \bf
EARL: Eye-on-Hand Reinforcement Learner for Dynamic Grasping \\ with Active Pose Estimation
}

\author{Baichuan Huang,  Jingjin Yu, and Siddarth Jain % <-this % stops a space
\thanks{Mitsubishi Electric Research Laboratories (MERL), Cambridge, MA 02139, USA {\tt\small baichuan.huang@rutgers.edu, jingjin.yu@rutgers.edu, sjain@merl.com}} %Author Baichuan Huang was partially supported by NSF IIS-831079.}
}

\begin{document}

\maketitle
\thispagestyle{empty}
\pagestyle{empty}

%%%%%%%%%%%%%%%%%%%%%%%%%%%%%%%%%%%%%%%%%%%%%%%%%%%%%%%%%%%%%%%%%%%%%%%%%%%%%%%%
\begin{abstract}
In this paper, we explore the dynamic grasping of moving objects through active pose tracking and reinforcement learning for hand-eye coordination systems. Most existing vision-based robotic grasping methods implicitly assume target objects are stationary or moving predictably. Performing grasping of unpredictably moving objects presents a unique set of challenges. For example, a pre-computed robust grasp can become unreachable or unstable as the target object moves, and motion planning must also be adaptive. In this work, we present a new approach, \b{E}ye-on-h\b{A}nd \b{R}einforcement \b{L}earner (\b{\earl}), for enabling coupled Eye-on-Hand (EoH) robotic manipulation systems to perform
real-time active pose tracking and dynamic grasping of novel objects without explicit motion prediction. \earl readily addresses many thorny issues in automated hand-eye coordination, including fast-tracking of 6D object pose from vision, learning control policy for a robotic arm to track a moving object while keeping the object in the camera’s field of view,
and performing dynamic grasping. We demonstrate the effectiveness of our approach in extensive experiments validated on multiple commercial robotic arms in both simulations and complex real-world tasks. 

\end{abstract}

%%%%%%%%%%%%%%%%%%%%%%%%%%%%%%%%%%%%%%%%%%%%%%%%%%%%%%%%%%%%%%%%%%%%%%%%%%%%%%%%
%\input{sections/intro}
\section{INTRODUCTION}
\label{sec:intro}

% \begin{figure}[t!]
% \vspace{1mm}
%     \centering
%     \includegraphics[width = 0.98\linewidth]{figures/system.pdf}
%     \caption{\textit{Top:} An example of an Eye-on-Hand (EoH) system used in this research, with a wrist-mounted RGB-D camera and a Universal Robots UR-5e arm. \textit{Bottom:} Sample of experimental tasks used to evaluate \earl for dynamic grasping in challenging setups. \emph{Extended workspace} shows that the tracking works outside the workspace in which the system is trained. \emph{Over and behind wall} shows the system tracks the object smoothly as it is moved across a barrier. \emph{Placing on box} shows the system tracks a novel object with significant elevation change.  
%     \label{fig:system-snapshot}
%     }
%     \vspace{-2mm}
% \end{figure}

\begin{figure}[t!]
\vspace{1mm}
    \centering
    % \includesvg[inkscapelatex=false, width = 0.98\linewidth]{figures/system.svg}
    \includegraphics[width = 0.98\linewidth]{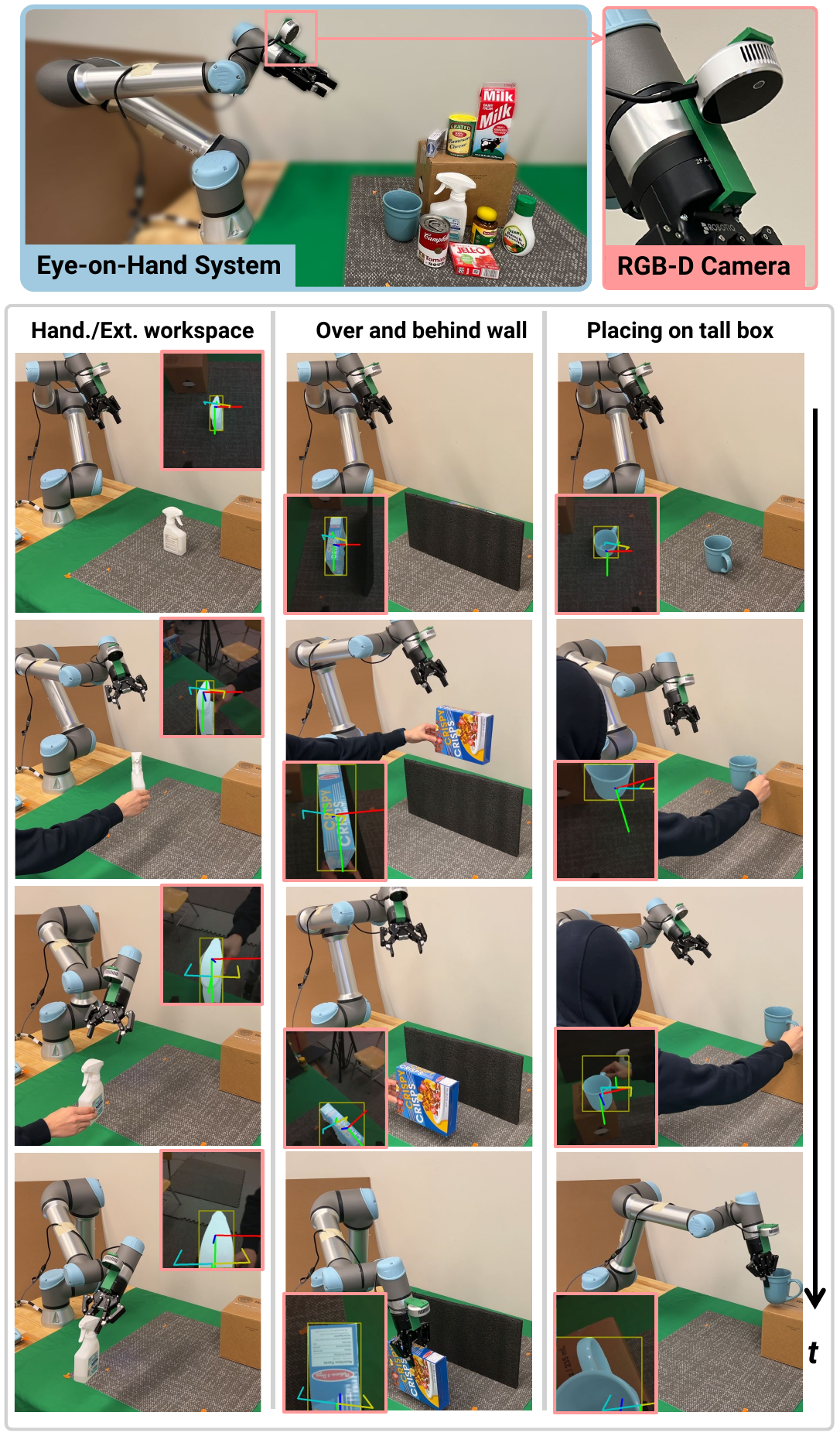}
    \vspace{0.5mm}
    \caption{\textit{Top:} Eye-on-Hand (EoH) system with a wrist-mounted RGB-D camera and a robotic manipulator. \textit{Bottom:} Sample of experimental tasks used to evaluate \earl in challenging setups. \emph{Hand./ext. workspace} shows that the system works outside the trained workspace and can perform human-robot handover. \emph{Over and behind wall} shows the system can overcome occlusions typical external camera setup faces by tracking across a barrier. \emph{Placing on box} shows the system can handle significant elevation changes. 
    \label{fig:system-snapshot}
    }
    \vspace{-2.5mm}
\end{figure}

Robotic manipulation of everyday objects in dynamic environments constitutes a \emph{fundamental skill} for enabling the next generation of advanced robotic systems. Providing robots with 6 degrees of freedom (DoF) pose tracking and grasping capability in unstructured and dynamic environments beyond static tabletop scenarios can benefit many automation applications. For example, a human handing over an object to the robot, assembly of industrial parts, etc. 

Autonomous grasping and manipulation of objects in stationary settings have been studied extensively~\cite{kleeberger2020survey}. Dynamic environments bring many challenges for performing grasping. First, the target object might move with an unknown motion, which requires understanding and predicting the object's motion or continuous tracking and active following by the manipulator. Second, computed motion plans can become obsolete, and thus dynamic environments require online or fast replanning. Additionally, the approach direction for grasp planning changes with the object's motion, and therefore, a stable grasp can become unreachable and unstable as the target object moves. Lastly, active perception~\cite{chen2011active} is required for dynamic grasping with Eye-on-Hand (EoH) systems, as the robot can lose track of the target object because of the robot's motion or as the target moves away from the field of view (FoV) of the camera. 
 
A supermajority of vision-based manipulation systems typically utilize cameras fixated above the workspace \cite{huang2022parallel, tuscher2021deep}. Thus, the perception subsystem (e.g., RGB-D cameras) and the manipulation system (e.g., robot arms) are \emph{decoupled}. Such settings implicitly assume ideal viewing distances and angles for focusing on target objects. This fixation may require large clearances above/around the workspace, rendering the overall system inflexible and unsuitable for some applications, especially when occlusions are unvoidable or spaces are confined, e.g., retrieving a condiment jar in a cabinet or performing an inspection in a pipe using a snake-like robot \cite{yamada2013snake} and robot dog. Such decoupling can cause occlusions or loss of tracking in many scenarios, especially when the target object is moving. Thus, mounting a camera on the wrist of a manipulator can be an alternative~\cite{feng2019robot}. In some cases, the vision sensor is mounted on the robot arm near the end-effector \cite{marturi2019dynamic, wu2022grasparl}; however, pose tracking has been carried out afar with another static workspace camera, rendering the perception process decoupled from the object manipulation operation. 

In this work, we relax some of the assumptions and address the challenges mentioned above by tackling robotic grasping with an ``Eye-on-Hand'' (EoH) system, where the sensory perception system is coupled with the manipulator (see. Fig.~\ref{fig:system-snapshot}, \textit{Top}). The binding constraint on the ``eye" and the ``hand" subsystems, dynamic range changes for perception, generalization to different objects, and no prior knowledge of the object's motion profile bring many challenges. We propose a novel solution based on \emph{active pose tracking and reinforcement learning}, which enables EoH robotic systems to execute dynamic grasping of moving objects in real-time. Our proposed method, called \textbf{E}ye-on-h\textbf{A}nd \textbf{R}einforcement \textbf{L}earner (\earl), approaches the target with the constraint of keeping the target object in the FoV, despite drastically changing distances and view angles between the camera and the target object. We propose a reinforcement learner that maps from the moving object pose differentials and adaptive changes of grasp pose to desirable robot arm joint velocities for dynamic grasping. We propose to perform active pose tracking to encode visual observations and perform the training through a carefully designed curriculum to overcome the challenges. %Our proposed method can handle tracking a wide range of objects with deep-learned features and pose graph optimization methods. 
Through extensive evaluation, we demonstrate that \earl robustly and effectively tracks and approaches the moving target object until successful grasping is realized.

In summary, our work makes the following contributions:
\begin{itemize}[leftmargin=3.5mm]
%\vspace{-2mm}
    \item It introduces a high-performance manipulation framework for Eye-on-Hand (EoH) robotic systems consisting of proposed active pose tracking with a moving camera and a curriculum-trained reinforcement learning method \earl with sim-to-real generalization capability for full 6-DoF dynamic grasping of novel objects with no prior knowledge of the object's motion profiles. 
    \item  It extensively evaluates our proposed framework on a set of novel objects with different motion types and complex tasks in various settings for dynamic grasping in simulation and the real world, including multiple robotic arms as EoH systems. 
\end{itemize}
%\vspace{-1mm}

We explore related work in Sec.~\ref{sec:related}. System setup and task description are described in Sec.~\ref{sec:system}. In Sec.~\ref{sec:technical}, we present the overall framework \earl and our proposed method. Experimental details are described in Sec.~\ref{sec:experiments}. We discuss and conclude in Sec.~\ref{sec:conclusion}.

\section{Related Work}
 \label{sec:related}

Vision-based robotic grasping solutions~\cite{du2021vision} can categorize along several dimensions. Model-based approaches rely upon knowledge about the target object~\cite{wan2020planning}, e.g., a CAD model. Model-free methods directly propose grasp candidates and aim for generalization to novel objects. Analytic or geometric methods analyze the shape of a target object for grasp pose identification~\cite{jain2016grasp}. Data-driven grasp synthesis uses learning~\cite{kleeberger2020survey} and has significantly progressed due to better learning methods and data availability. Some approaches sample and rank grasp candidates using deep neural networks~\cite{liang2019pointnetgpd}. 
Reinforcement Learning (RL) approaches also find applicability for grasp synthesis~\cite{mohammed2020review, zhang2023cherry}, where suitable candidates are learned and validated with repeated interactions in a simulation or real environment. 

Another essential distinction is open or closed-loop grasping. Approaches performing open-loop grasp execution are sensitive to calibration errors, failing to handle dynamic environments. Some methods tackle closed-loop grasping using continuous visual observations~\cite{morrison2018closing} with visual servoing control~\cite{viereck2017learning} or RL. Some methods require vast data for training; for example, QT-Opt~\cite{kalashnikov2018scalable} collected data within several weeks across seven robots. Many methods characterize by constrained state-action spaces, for example, focusing on top-down grasping~\cite{kalashnikov2018scalable, morrison2018closing, burgess2022eyes} in 4-DoF, and thus are limited in task scope. Only a handful of methods target grasp synthesis for 6-DoF~\cite{song2020grasping} to grasp objects from a broader range of unstructured settings beyond the tabletop scenario. 

Most existing methods focus on static environments, typically utilizing cameras fixed in the workspace to perform grasp synthesis. Such settings can suffer occlusions in many scenarios, especially when the robot is approaching the target. Employing multiple cameras can help mitigate occlusion in some scenarios. In real-world tasks, interacting with dynamic objects or manipulating objects while a robot is in motion can benefit many applications. For such dynamic grasping tasks, mounting a camera on a robot's wrist can significantly reduce occlusion's impact and enable the robot to perform grasp synthesis in various scenarios. 

Grasping in dynamic environments presents additional challenges and requires the robot's motion and grasp planning to be adaptive and real-time. Morrison et al.~\cite{morrison2018closing} present a generative grasping convolutional neural network (GG-CNN) for fast closed-loop grasping in slightly moving scenes. Their method generates quality measure for every pixel in depth images. Similar to our work, they use a wrist-mounted camera but only evaluate their approach for 4-DoF top-down grasping. Moreover, GG-CNN has high requirements for data annotation. Marturi et al.~\cite{marturi2019dynamic} dynamically plan trajectories to grasp a moving object based on visual information. However, their approach uses multiple cameras. Song et al.~\cite{song2020grasping} propose an end-to-end RL method for closed-loop grasping where a DNN models a value function that maps the images from the wrist-mounted camera to the expected rewards in that state. The approach can handle 6-DoF grasps for slightly moving scenes but only operates on a discrete action set. One common approach for dynamic grasping involves motion prediction for the target object to improve success. The motion prediction ability can be helpful in both planning a grasp and approaching the object. Akinola et al.~\cite{akinola2021dynamic} introduce a reachability and motion awareness solution. They implement a recurrent neural network (RNN) for modeling and predicting object motion that works well for linear, sinusoidal, and circle trajectories. However, it could be of poor generalization in unseen and complex trajectories, e.g., random movements, and their solution uses a position-fixed camera. In this work, we propose a model-free RL-based method with a wrist-mounted camera for picking a moving object with continuous actions while keeping the object in the FoV of the robot. Our approach does not limit the grasp synthesis to a top-down direction and generalizes well for unseen and complex trajectories.
%without explicit modeling of motion prediction.
%\input{sections/robot_hardware}
\section{System Setup and Problem Description}
\label{sec:system}

%\subsection{System: Coupled Eye-on-Hand (EoH) Manipulator}\label{subsec:eoh}
This work endows Eye-on-Hand (EoH) systems with the capability to track and grasp moving target objects. The EoH system comprises a high-DOF robotic manipulator, an end-effector, and a wrist-mounted sensory perception system that is fixated on the manipulator near the end-effector. The coupling between the perception and manipulation subsystems means that such an EoH system can operate without workspace constraints and occlusions faced by most existing systems \cite{huang2022interleaving} using an external camera.

We instantiate experiments and demonstrations on two robotic arms as such EoH systems to showcase our methods' generality. These systems use Universal UR-5e (6-DOF) and Kinova Gen3 (7-DOF), respectively, both commanded using \emph{joint velocity control}. The vision system includes an Intel RealSense L515 RGB-D camera for perception, secured on the gripper using a custom 3D-printed mount. We use a Robotiq 2F-85 two-finger gripper as the end-effector. 

\noindent \textbf{Problem Definition:} We focus on enabling coupled EoH systems to perform dynamic grasping in 6-DoF (i.e., $SE(3)$) of a moving object with \emph{a priori} unknown motion. We make no assumptions about the shape or identity of target objects other than that they are rigid bodies and graspable by the end-effector. Additionally, the target object can move freely in the robot's reachable workspace (approachable by the robot). Completing this task with the EoH system requires another sub-task of tracking the moving object, following it with the robot's motion such that it keeps the target object in the FoV of the EoH camera while approaching the target. The task is successful if the robot can grasp and pick up the object.

\begin{comment}
Our primary focus in this work is to enable coupled EoH systems described in Sec.~\ref{subsec:eoh} to per from dynamic grasping in 6D (i.e., $SE(3)$) while tracking  moving object and keeping the target in view of the eye-in-hand camera while approaching the target. Given a target object, the proposed perception system continuously tracks it through motion until the object can be grasped with a good level of certainty with the EoH system. A grasping action is attempted autonomously by our system while tracking the object opportunely when the target is within the end-effector reach. Note that we make no specific assumptions about the target objects other than that they are rigid bodies (for pose tracking) and are graspable by the end-effector. Additionally, the target object can move freely in the manipulator's reachable workspace at reasonable speeds (approachable by the robot). 
\end{comment}

\section{\earl Framework}
\label{sec:technical}

\begin{figure*}[t!]
% \vspace{1.5mm}
    \centering
    % \includesvg[inkscapelatex=false, width = 0.98\linewidth]{figures/framework.svg}
    \includegraphics[width = 0.98\linewidth]{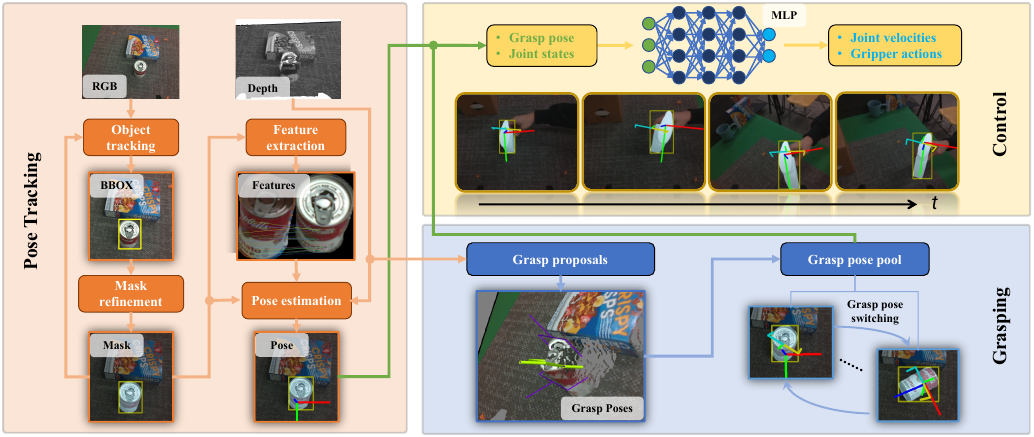}
    \caption{Illustration of the proposed framework. RGB-D image input initializes object tracking and mask refinement for pose estimation components. The pose tracking module provides a 6D pose of the target object. A grasp pose pool is built from a grasp proposal network. The grasp planning module computes a suitable grasp pose on the target. The control policy commands the EoH system to track and finally grasp the target using the robot arm. 
    \label{fig:method-overview}
    }
    % \vspace{-8mm}
\end{figure*}

In this section, we describe the various components of our framework (illustrated in Figure~\ref{fig:method-overview}). First, we present our framework overview. We then describe our visual processing unit that tracks target object poses in real-time. We then describe the grasp planning pipeline that continuously adapts grasp selection. Finally, we present our arm motion generation method based on curriculum-trained RL. 

\subsection{Framework Overview}

Grasping in a dynamic environment presents many challenges. A stable and reachable grasp can become unreachable and unstable as the target object moves. The EoH system can also lose track of the dynamic target object. We employ RL for EoH systems, where we learn a control policy (\earl) in simulations for dynamic grasping of arbitrary objects, and we propose to utilize the object's active pose information as visual feedback. Our proposed method enables dynamic grasping with tracking of novel objects without explicit motion prediction requirements. First, we perform pose estimation of the target with the moving EoH camera for providing visual feedback. We select a design that continuously tracks the target object in 2D using learned features~\cite{mayer2022transforming,yan2021alpha} and subsequently performs pose estimation using a combination of 2D features~\cite{r2d2} and depth information~\cite{wen2021bundletrack} to speed up the processing and ensure sufficient tracking accuracy. The control policy is a Proximal Policy Optimization (PPO) based~\cite{schulman2017proximal} RL framework which maps object pose differentials to desirable robot arm joint velocities. Since we work with an EoH system, the robot arm must continuously adjust the pose of the RGB-D camera to keep the target object in the FoV. Simultaneously, the control policy must guide the end-effector to approach the object and grasp it. Achieving all these requirements naturally and smoothly in real-time is only possible through a finely-tuned, multi-stage control policy, as described in Sec.~\ref{sec:rl}. We perform grasp planning with multiple grasp proposals on the target object. The best grasp pose is dynamically updated, which may happen as the target object continuously moves in 6D. We dynamically select a suitable candidate for biasing the control policy actions. The overall framework is shown in Fig.~\ref{fig:method-overview}, and the steps are detailed in the following sections.

\subsection{Active Object Pose Tracking}\label{subsec:pose-tracking}
% \r{TODO: discuss what is needed for real-time tracking of moving objects with a moving camera, and our choice of subsystem design.}
%We integrated four previous works as our single object pose tracking system, as shown in Fig.~\ref{fig:method-overview}.
%
%Some works do the object pose estimation without CAD models, like OnePose~\cite{sun2022onepose}, but they require a scan of the object at the start.
%
%However, the robot should start tracking immediately after the first frame, where a human operator specifies the target in our setting.
%
%In this paper, the pose tracking module provides an accurate and consistent pose of the target object given a series RGB-D image, where both the camera and object could move.
%

Active perception~\cite{chen2011active} implies computer vision implemented with a movable camera, which can intelligently alter the viewpoint to improve the system's performance. In this work, we consider active tracking with an Eye-on-Hand camera. Many methods focus on top-down grasping~\cite{kalashnikov2018scalable, morrison2018closing}, where they constraint the target object in the 2D workspace. This case can be relatively simple. In this work, we target grasp synthesis in 6-DoF, which requires the robot to be aware of the target object's 3D position and 3D orientation. We perform 6D pose estimation with the EoH system for tracking and computing an encoding of the target object from visual observations. Typically, 6D object pose estimation methods~\cite{du2021vision} assume known object models and can categorize into correspondence-based, template-based, and voting-based. We make no assumptions about the object's model and motion profile and continuously track the object's pose with the moving camera. 

Given the first frame RGB-D image $I_0$ containing the target object $O$, we aim to continuously track $O$'s 6D pose relative to the camera at any time $t$ in image $I_t$. We realize this by a correspondence-based approach and leverage BundleTrack~\cite{wen2021bundletrack} method for maintaining a keyframe memory pool and performing online pose graph optimization that takes both feature correspondences and dense pixel-wise correspondences from a depth image. To accomplish fast and active pose tracking of novel objects in motion using only a sequence of RGB-D images, we introduce several augmentations to enhance the BundleTrack. 

At time $t$, the pose tracking process starts with obtaining a bounding box $B_t$ of the target object, employing a transformer-based 2D object tracker~\cite{mayer2022transforming}, capable of tracking an object through multiple frames with a series of RGB images. We then enhance the fit of the bounding box $B_t$ with Alpha-Refine~\cite{yan2021alpha} and simultaneously obtain an object mask $M_t$. Alpha-Refine~\cite{yan2021alpha} is a boosting method that extracts and maintains detailed spatial information. These augmentations support fast tracking and work with novel objects without requiring additional training or object models. We utilize the mask $M_t$ to provide more accurate RGB and depth information as input to the BundleTrack, leading to direct performance gains for fast-tracking the 6D pose. A key component for correspondence-based pose estimation is the feature extractor. We employ R2D2~\cite{r2d2} feature detector, which predicts a set of sparse locations as object features in the input image that are repeatable and reliable for local feature matching and thus outperforms the LF-Net~\cite{ono2018lf} detector utilized in the BundleTrack. We note that the color image input to R2D2 contains only the target object cropped based on $M_t$. Our method enables more robust active pose tracking in real-time robot control.

\subsection{Grasp Planning}
% \r{TODO: discuss system initialization, including grasp proposals}

We use the first RGB-D image to initialize the pose tracking (see Sec.~\ref{subsec:pose-tracking}) and compute grasp proposals.
Selecting a good set of candidate grasp pose is important for two reasons. First, computing grasp poses on the fly is time-consuming, making it difficult to achieve real-time performance. Second, it is required to compute a desired output from a control policy towards a graspable pose on the moving object that drives the end-effector close to the target object. For grasp planning, we compute a set of grasp poses using Contact-GraspNet~\cite{sundermeyer2021contact}, which utilizes a raw depth image and object mask as input to generate a set of 6-DoF grasp proposals. This approach works on novel objects and environments without the need for fine-tuning. Next, we compute a set of transformations from grasp poses to the 6D pose of the target object and store them in a grasp memory pool. All proposed grasp poses are relative to the target object's pose, which helps us avoid doing computationally expensive real-time grasp synthesis. As both the EoH and the object move, the system dynamically chooses which grasp pose to track from the grasp memory pool at a given time. The selection of the grasp pose is driven mainly by \emph{angular distance}. We use quaternion representation to store the orientation information. At each control iteration, the orientation of gripper $q_g$ is compared with that of all grasp poses $q_i \in G$ via 
$$
\Delta q_i = q_g * q_i^{-1} , 
d_i = 2\arcsin(\min(\|\Delta q_i\|, 1)).
$$
The grasp pose with the smallest $d_i$ is set to be the current target pose to be tracked.

\subsection{Coupled Control of EoH System}\label{subsec:pd}

Given active pose tracking and grasp planning for adjusting the target grasp for a moving object, next, the EoH system must track and follow the target and successfully perform dynamic grasping. Two constraints must be jointly satisfied by a policy to control the robot's approach and grasp the target object: maintaining the object in the camera's FoV and avoiding collisions, including self-collision. 

As a first solution, we develop a standard PD controller driven by an inverse kinematics (IK) solver for the control policy. We treat this solution as a baseline approach for performing dynamic grasping with the EoH system. 

\begin{wrapfigure}[16]{r}{0.46\linewidth}
    \centering
    % \includesvg[inkscapelatex=false, width=\linewidth]{figures/baseline.svg}
    \includegraphics[width = \linewidth]{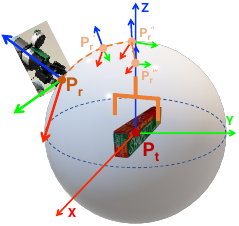}
    \caption{Baseline control policy. Target grasp pose on the object is shown in orange. The robot arm should follow the trajectory of $P_r^{'}, P_r^{''}, P_r^{'''}$ to reach the target object and finally grasp it at $P_t$. %\bh{I should construct this in 3D and then render in 2D. The perspective looks unreal}
    }
    \label{fig:baseline}
    % \vspace{-8mm}
\end{wrapfigure}

\vspace{1mm}
\noindent \textbf{Baseline control policy:} Given an end-effector goal as a grasp pose target to the EoH system at every frame, using the IK (damped least squares as described in ~\cite{buss2004introduction}) we develop a servoing method which generates desired joint velocity commands using a PD control formulation. At a higher level, the servoing-based baseline repeatedly computes the pose of the next goal for the end-effector following a hand-designed trajectory optimization, shown in Fig.~\ref{fig:baseline}. 

The EoH system navigates the gripper (with pose $P_r$) ``above'' the object to a pose at some distance from the object, from where a straight-line movement of the gripper can lead to the desired grasp pose $P_t$. In the figure, this stage is achieved as $P_r$'s $z$-axis aligns with the object pose's $z$-axis. To realize the design, a pose waypoint for the gripper $P_{r}^{'}$ can be computed as position $P_{r.p}^{'}$ and rotation $P_{r.q}^{'}$. The next position of pose $P_{r.p}$ needs to move the gripper toward the approaching direction ($z$-axis of $P_t$) while maintaining a fixed distance, 
$$r=|P_{r.p} - P_{t.p}|=|P_{r.p}^{'} - P_{t.p}|.$$ 
We realize this by moving on a sphere with a radius of $r$ as shown in Fig.~\ref{fig:baseline}. If the $z$-axis between $P_r$ and $P_t$ are well aligned (smaller than a threshold $\theta_{align}$, set to $0.175\si{\rad}$), then, $r$ is decreased, otherwise, $r$ is increased without exceeding $r_{\max}$ (set to $30\si{\cm}$). 

%the gripper pose should move along the path $P_r \rightarrow P_r^{'} \rightarrow P_r^{''} \rightarrow \cdots \rightarrow P_t$ as shown in the Fig.~\ref{fig:baseline}.
%
%The $P_{r.q}^{'}$ is then determined by $P_{t.p}$ and $P_{r.p}^{'}$, where the $z$-axis of the next gripper pose is the vector $P_{r.p}^{'} - P_{t.p}$, and $y$-axis is set to be the same one as $P_t$ with some interpolation, lastly $x$-axis is known as we are using right-hand coordination system.

\section{Training \earl}\label{sec:rl}

% \TODO{Alternatively, Control Policy Design for Coupled EoH System, if we put the PD control here as well.}
%\TODO{what is the basic idea of using RL and training it effectively for this task. What is the novelty?}

%The baseline control policy can work in practice; however, it may produce jerky or unsafe behavior and may not ensure a high success rate. As a more robust solution, 

We propose an Eye-on-Hand reinforcement learner (\earl) as a control policy that directly maps the desired grasp pose and joint states to the robot's desired joint velocity and gripper actions. Training an RL model with many continuous DoFs is challenging due to long policy horizons and sparse reward signals. In our case, the challenge is compounded by the strong coupling between vision and manipulation subsystems. To effectively train \earl in simulations, we formulate the task as a single-agent Markov Game. In each episode, a target object (possibly among other objects) is randomly placed in the workspace. The target object may randomly translate and rotate in the 3D workspace (gravity is disabled so that the object can move in the air). The robot is tasked to grasp the target object using a specific grasp pose. The environment resets itself upon a grasp success/failure or if object tracking fails (out-of-view). 

%
%The robot arm is fixed at the origin and does joint velocity control.
%Differently, the robot gripper (two-finger Robotiq 2F-85) does position control.
%We considered 6 DoF (UR5e) and 7 DoF (Kinova-Kortex) arms in our task.
%

%Most RL algorithm requires a long training time, as collecting data from the simulator takes time.
%

%\r{We should have a figure somewhere in this section showing the Isaac environment, which also shows the curriculum training feature of \earl.}

\begin{figure}[h]
    \centering
    % \includesvg[inkscapelatex=false, width = 0.95\linewidth]{figures/rl.svg}
    \includegraphics[width = 0.95\linewidth]{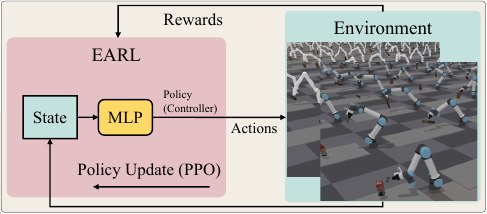}
    \caption{Overview of training EARL.
    \label{fig:rl}
    }
    % \vspace{-3mm}
\end{figure}

\vspace{1mm}
\noindent \textbf{Choice and Design of Learning Framework.}
%We adopt PPO~\cite{Schulman2017ProximalPO} to train \earl as RL Games~\cite{rl-games2022}, which can be performed in an end-to-end fashion in Isaac Gym. In our case, the MLP for PPO has three hidden layers of size $[256, 128, 64]$. 
%
We train \earl policy with a PPO~\cite{schulman2017proximal} agent, which receives the state quantities of the target object and the EoH system and decides which action will be taken to execute dynamic grasping. We employ Isaac Gym~\cite{makoviychuk2021isaac} for efficient RL and data acquisition, capitalizing on PPO's inherent support for parallel training. We perform simultaneous simulations of thousands of independent EoH systems in Isaac Gym for model training. The overview of training EARL is shown in Fig.~\ref{fig:rl}. The MLP for PPO has three hidden layers of size $[256, 128, 64]$. 

For input, the reinforcement learner needs information on the target object, which we provide as an encoding represented by the object's pose (translations and orientation), in addition to the state of the robotic arm. We describe a unique pose representation for the input. Because Euler angles have singularities and quaternions are non-unique, training the RL agent with such parameterizations can be difficult in~\cite{zhou2019continuity}. Borrowing from~\cite{allshire2022transferring, narang2022factory}, we use four \emph{keypoints} to represent the translation and orientation of a rigid body. These keypoints can be represented in many ways, and in our settings, we describe them using the corners of a fixed size square where the square is selected to coincide with the $XY$ plane of the coordinate system of the given object's pose, with its center coinciding with the origin of the pose coordinates.  We represent the input as, $$S=[K_r, K_{rg}, K_{gg'}, J_p, J_v, c],$$
where $K_r$ are the keypoints of the gripper. Let $K_g$ be the target (grasp) keypoints, then $K_{rg} = K_r - K_g$ is the error that should be minimized (reach $0$) for a successful grasp. $K_{gg'} = K_g - K_{g'}$ is the change of the target grasp keypoints between the current and previous frames. $J_p$ and $J_v$ are the robot arm's joint position and velocity vectors, respectively. $J_p$ is scaled to $[0, 1]$ depending on the rotation limits. $c$, a Boolean variable, indicates whether the gripper is closed or open. Policy actions, as the output of the MLP, are the velocity target for each robot's joint and a boolean command for the gripper. We choose joint velocities to control the robot arm because it results in comparatively smoother motions than position-based control. 

\vspace{1mm}
\noindent \textbf{Reward Shaping.}
The learning process can be much faster and successful through a suitable reward schedule with diverse guidance. We carefully perform reward shaping to encourage the robot to grasp the target object fast and securely. On a high level, the problem can be seen as tracking and grasping. The tracking portion moves the end-effector closer to the desired grasp pose while maintaining the camera's view of the object and simultaneously avoiding collisions. The grasping component needs to determine an opportune moment for closing the gripper. We consider the case where the motion of the target object can be unpredictable, so the best strategy for the robot arm is to continuously track the target and grasp it at an opportune moment. We design the reward to have multiple components, 
$$R = P_{dist} + P_{view} + P_{coll} + P_{over} + P_{alig} + P_{move} + R_{grasp}.$$

$P_{dist}$, a dense penalty, is the distance between the gripper pose and grasp pose, which encourages the robot to minimize $K_{rg}$.
$P_{view}$ is a penalty given when the target pose is out-of-view. This penalty  helps the active pose estimation with the moving camera by keeping the object in FoV.   
$P_{coll}$ is a penalty for any collision before grasping. 
It is easy for the robot to keep the object in view and avoid collision by staying far from the target object, a behavior that must be avoided. 
For that, $P_{over}$ penalizes the agent when the distance between the gripper and grasp pose is larger than a threshold. $P_{alig}$ is a helper to regularize arm motion and help with learning. It contains three parts, the z-axis alignment of the gripper and grasp pose, the y-axis alignment of the gripper and grasp pose, and the centering of the object in the camera view. The z-axis and y-axis alignments guide the robot's approach to the target. The closer the gripper is to the grasp pose, the higher probability the target object may collide with the robot or move out of the FoV, as the target object could randomly move. A $P_{move}$ penalty is introduced to keep the robot's gripper $\theta_{d}$ (set to $20\si{\cm}$) away from the object if the speed of the object is faster than $\nu_{o}$ (set to $4 \si[per-mode=symbol]{\cm\per\s}$). $R_{grasp}$ is the reward for closing the gripper when the target object has been successfully grasped. We use the contact force of two fingers to determine whether the gripper has grasped the target object in the simulation.

\noindent \textbf{Curriculum Design.}
Learning dynamic grasping can be challenging for EoH systems, and training such a network from scratch with many continuous DoFs is challenging. We explore a curriculum design to train the EoH system efficiently. In particular, we use a three-stage curriculum~\cite{narvekar2020curriculum}, gradually increasing the task's difficulty and dynamically changing the rewards scalar for efficient learning. In the first stage, the object is randomly placed in the workspace, and the robot's gripper is randomly sampled closer toward the grasp pose on the object. The object randomly moves at a low speed ($<=\nu_{o} \si[per-mode=symbol]{\cm\per\s}$). In addition, an episode is not terminated when the object is out-of-view. These relaxations significantly limit the initial search space to help training. Once the success rate of the initial policy is over a threshold, in the second stage of training, the environment will reset once the object is out-of-view, denoting task failure. The penalty for $P_{view}, P_{coll}, P_{over}, P_{move}$ are increased, and they will continue to increase to reinforce that the robot should avoid these unwanted scenarios. %These penalty scales are set to be low at the start since a high penalty causes the policy to be passive. 
Low penalty causes the policy to be aggressive, which leads to high collision rates or not keeping the object in view. On the other hand, a high penalty discourages the robot from approaching the object. We dynamically change penalty scales to encourage different behaviors and achieve a high success rate. In the last stage, we let the object move faster (the max $\nu_{o}$ is set to $8.5\si[per-mode=symbol]{\cm\per\s}$), and the robot gripper is fixed far from the target object at the start of each episode, working in the full workspace. 

The RL training parameters are identical for UR-5e (6-Dof) and Kinova-Gen3 robots (7-Dof). The only difference is that the penalty scales of $P_{view}$ and $P_{coll}$ are set to be lower for Kinova due to the robot's kinematics, which has long links between joints, making it comparatively harder to train. Once trained, our policy \earl runs in real-time on new objects and the training time amounts to just $\sim$4 hours of the learning experience.

\vspace{1mm}
\noindent \textbf{Enabling Direct Sim-to-Real Adaption.}
A goal of \earl is to have RL agents trained in a simulator directly applicable to real-world EoH systems. We developed an effective technique of independent interest based on an observation that the sim-to-real gap does not strongly correlates with a control policy, and we encode visual observations as high-level representations. Instead, the gap is mainly caused by parameter differences between the simulation and the real system. With this observation, we \emph{decouple} the sim-to-real gap reduction from training \earl. Using only the baseline PD-based controller (Sec.~\ref{subsec:pd}), running in both simulation and real-world EoH systems, we fine-tuned the necessary parameters to reduce the sim-to-real gap. For example, because real robot arms are torque-controlled at a lower level, damping parameters must be appropriately adjusted to realize accurate joint velocity control. 

%
\begin{comment}
We have tried to reduce the sim-to-real gap by implementing a baseline method that does not require training.
%
We compare the performance of the baseline method in simulation and the real robot, tuning the damping parameters of the robot in the simulator to match the real robot.
%
Underlying the robot joint control in the simulator and the real robot is a torque control.
%
The damping defines how much torque should be applied to reduce the difference between the target velocity and the current velocity of a joint.
%
With a reasonable damping parameter, the sim-to-real transfer becomes feasible.
%
Control frequency is also matched between the simulation and the real robot.
%
The pose tracking modules run at an average of 10 FPS, which is the upper bound of how frequently we can control the robot.
%
If the robot is trained with a higher frequency, the transfer to an actual robot will make the motion jitter.
\end{comment}

\section{EXPERIMENTS \& RESULTS}\label{sec:experiments}
\begin{figure}[t]
    \centering
    % \includesvg[inkscapelatex=false, width = 0.95\linewidth]{figures/object-s.svg}
    \includegraphics[width = 0.95\linewidth]{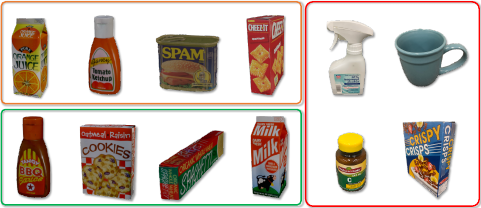}
    \caption{\textit{Top Left:} Training set objects. \textit{Bottom Left:} Validation set objects used in simulation. \textit{Right:} Real test set objects. Note that the real objects are significantly more challenging.
    \label{fig:objects}
    }
    %\vspace{-2mm}
\end{figure}

We evaluated the proposed methods both in simulation (Issac Gym~\cite{makoviychuk2021isaac, narang2022factory}) and on real robots (Universal UR-5e 6-DoF (R1) and Kinova Gen3 7-DoF (R2)) as EoH systems. Both robots are equipped with a two-finger Robotq 2F-85 gripper and an Intel RealSense L515 camera. The workspace for training is a cubic region of $40 \times 40 \times 40 \si[per-mode=symbol]{\centi\meter\cubed}$. We evaluated our system on machines with a single GPU (Nvidia 3090, uses 8 GB memory). Our method can handle novel objects for which grasp pose can be reasonably tracked, allowing us only to use a few objects (four) for training. We tested in simulation and the real world with four unseen sets of objects each. Objects are selected from the YCB and HOPE datasets~\cite{calli2015ycb,tyree20226}; and some household items. Fig.~\ref{fig:objects} shows these objects. The pipeline runs at 15 FPS.
%
%The implementation of our framework will be made available at \url{github}.

\subsection{Simulation Experiments}

\begin{figure}[h]
% \vspace{1.5mm}
    \centering
    % \includesvg[inkscapelatex=false, width = 0.9\linewidth]{figures/motion.svg}
    \includegraphics[width = 0.9\linewidth]{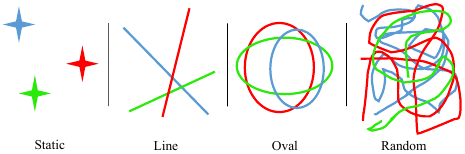}
    \vspace{1mm}
    \caption{Four motions are considered in the simulation evaluation. The figure illustrates these motions in 2D for visualization purposes, which happen in 3D and 6D. The RL control policy is trained only for random (6D) motion. 
    \label{fig:motion}
    }
    % \vspace{-8mm}
\end{figure}

\noindent \textbf{Task and Object Motion.}
The main objective of simulation experiments is to evaluate the control policies \mtf and \earl for the dynamic grasping problem. In the simulation experiments, similar to training, the simulator gives the object's pose, and we use one grasp per trial. To evaluate the policy, we created test cases that have four different motions as shown in Fig.~\ref{fig:motion}.

\begin{itemize}[leftmargin=3.5mm]
    \item \emph{Static} (S). Objects are randomly sampled in the workspace.
    \item \emph{Linear} (L). Line paths in $XYZ$ space are randomly sampled from boundary to boundary. 
    \item \emph{Oval} (O). Oval paths in the $XY$ plane are randomly sampled, and the objects stop moving at a random time step. The $z$-axis is also randomly sampled beforehand.
    \item \emph{Random} (R). The object randomly moves for some time, then stops moving for five seconds and repeats.
\end{itemize}

%The grasp pose of the target object is reachable when the object is finally stopped.
%
%When the target object is not moving, we still give it low vibration speeds to mimic the uncertainty from pose estimation in the real world, as we did in policy training. 
%
The object randomly moves with speed along any dimension set to a max. $5 \si[per-mode=symbol]{\cm\per\s}$ for translation, and $0.1 \si[per-mode=symbol]{\rad\per\s}$ for rotation. In the simulation, the target object never stops moving, even when slowed down considerably at random; it will ``vibrate'' around the center of mass to simulate uncertainty. 

\noindent \textbf{Metrics.}
We evaluate our method using the success rate, where the robot successfully grasped the target object in a given time (timeout). Three failure conditions are recorded: (1) collision, (2) losing track of the target object, and (3) the maximum episode length is reached.
%
%The collision considers the self-collision of robot, collision between robot and table, collision between robot and object while not close to the grasp pose.
%
%The rendering of an image is expensive in the simulator. Instead, we draw a vector from the camera center to the center of the target object and compare this vector to the \emph{lookat} vector of the camera. If the angular distance between these two vectors are greater than a threshold $21.8^{\circ}$ (the horizontal FoV of L515 is $56.4^{\circ}$), then it is considered as out of view.
Since active pose estimation is not performed in the simulation experiments, checking that an object is out of view is done by checking the object's pose center is beyond a certain threshold ($\sim 20^{\circ}$ in our experiments) away from the simulated camera's principle axis. 
%
%The maximum time per trial is 35 seconds, where the target object may randomly move in the first 30 seconds.
%The maximum time per trial is 35 seconds (timeout), where the target object may randomly move in that duration.
The maximum time per trial is 35 seconds (timeout). 
The robot is required to approach the target object and execute a swift grasp when appropriate.

\noindent \textbf{Results.}
We generated one thousand test cases in simulations across the motion types and test objects and repeated them for five trials. \earl control policy outperforms the baseline method on both robot systems R1 and R2, as shown in Table~\ref{tab:simtable}. The baseline method achieves an average success rate of $86.83\%$. \earl demonstrates an excellent average success rate of $95.73\%$. The primary failure mode of \earl is timeout due to the high penalty for collision and out-of-view, which encourages the robot to move conservatively toward the target object. Timeout failure for \earl is relatively low in random motion settings, likely due to the training being performed only using the case of random object motion. Nevertheless, \earl performs for other object motion types. In comparison, the performance of the baseline method goes down significantly as the difficulty of the task increases.

\begin{table}[h]
   % \lefting
   \vspace{2mm}
   \hspace{-1.5mm}
    \scalebox{0.96}{
    \begin{tabular}{rlcc|ccc}
    \toprule
        A$\vert$R & M & Success & Failure & Timeout & Collision & Out-of-View \\ 
        \specialrule{0.1pt}{2pt}{2pt}
        \multirow{4}{*}{B$\vert$R1} & S      & $94.90$ & $5.10$  & $0.20$ & $4.90$  & $0.00$\\
                                      & L   & $90.76$ & $9.24$  & $2.30$ & $6.54$  & $0.40$\\
                                      & O   & $86.54$ & $13.46$ & $1.14$ & $5.30$  & $7.02$\\
                                      & R & $76.28$ & $23.72$ & $5.34$ & $12.22$ & $6.16$\\ \specialrule{0.1pt}{2pt}{2pt}
        \multirow{4}{*}{E$\vert$R1} & S      & \textbf{95.76} & $4.24$ & $4.20$ & $0.00$ & $0.04$\\
                                      & L   & \textbf{95.00} & $5.00$ & $4.74$ & $0.00$ & $0.26$\\
                                      & O  & \textbf{96.50} & $3.50$ & $2.12$ & $0.70$ & $0.68$\\
                                      & R & \textbf{97.80} & $2.20$ & $0.60$ & $0.60$ & $1.00$\\ \specialrule{0.1pt}{2pt}{2pt}
        \multirow{4}{*}{B$\vert$R2} & S   & $95.90$ & $4.10$ & $2.32$  & $1.78$ & $0.00$\\
                                      & L & \textbf{95.08} & $4.92$ & $2.90$  & $1.78$ & $0.24$\\
                                      & O & $83.24$ & $16.76$ & $7.32$  & $3.86$ & $5.58$\\
                                      & R& $71.92$ & $28.08$ & $13.78$ & $7.80$ & $6.50$\\ \specialrule{0.1pt}{2pt}{2pt}
        \multirow{4}{*}{E$\vert$R2} & S & \textbf{96.72} & $3.28$ & $3.08$ & $0.00$ & $0.20$\\
                                      & L & $92.98$ & $7.02$ & $5.56$ & $0.50$ & $0.96$\\
                                      & O & \textbf{95.58} & $4.42$ & $2.32$ & $1.10$ & $1.00$\\
                                      & R & \textbf{95.48} & $4.52$ & $0.64$ & $3.18$ & $0.70$\\
    \bottomrule
    \end{tabular}
    }
    \vspace{1mm}   
    \caption{Simulation experiment results (all values in $\%$). A$\vert$R refers to the algorithm (baseline or \earl) and the robot 6-DoF (R1) or 7-DoF (R2). M refers to motion type (static (S), linear (L), oval (O), random (R)).}
    %One thousand cases (5 trials) on each movement style. The timeout is set to 35 seconds. Longer time does not change the statistics muc
    \label{tab:simtable}
\end{table}

%which can be attributed to the stiffness of the motion control using IK.
%
%he maximum joint velocity is set to be $0.2 \si[per-mode=symbol]{\rad\per\s}$ in both simulation and real robots.
%
%The IK solver computes the joint velocity, and we need to scale it to satisfy the maximum limit. However, the scaling is across all joints. Otherwise the ending gripper pose could be wrong to keep the object in camera's view or hit the object.
%
%The speed of robot arm is thus limited, whereas the RL policy learns to control each joint individually, and from our observation, the output joint velocity are often at $0.2 \si[per-mode=symbol]{\rad\per\s}$. This provides high reactive motion of arm when the target object moves in oval or random.
%
%Also, the IK solver does not consider self-collision, so the baseline collision rate is higher than the RL method.

\subsection{Real Robot Experiments}

\noindent \textbf{Tasks and Object Motion.}
The real experiments use actual perception and perform grasp planning with pose tracking by the EoH system. A human operator moves the object freely in the workspace based on the task scenario. We perform four motions (S/L/O/R) from simulation experiments to analyze the sim-to-real performance gap. The grasp pose pool is limited to one for a fair comparison. We also perform two more motion tasks for testing adaptive grasp switching: random and complex. The former executes random movements of the object, forcing a grasp pose switch to achieve success. The complex tasks involve a combination of motion types. These include three challenging tasks, as described in Fig. 1, which demonstrate generalization capability. 

\noindent \textbf{Results.}
We performed real robot trials using the UR-5e (6-DoF) as the EoH system for each method and motion type with each test object using random start and end positions (20 trials for each motion type per method). We ensure each trial is consistent across both methods. Table.~\ref{tab:realtable_ur5} show these results. \earl outperforms the baseline method for all test cases achieving an average success rate of $91.25\%$, versus $73.75\%$ with the baseline. Note that \earl policy is trained entirely in simulations. Comparing the simulation results demonstrates that we accomplish a small sim-to-real gap (less than 5\%) on the motions (S/L/O/R) with our sim-to-real adaptation. Furthermore, \earl also performed well on the random motion tasks and complex tasks with adaptive grasp switching, achieving an average success rate of $90\%$ and $80\%$, respectively. From our observation, the pose tracking module is bottleneck for achieving dynamic grasping without explicit modeling of motion prediction. We perform active pose tracking with object-level SLAM by considering texture, geometric information, and robust features; still, the RGB image suffers from motion blur and the depth image from noisy reflections and inaccurate measurements, affecting performance. Moreover, object tracking can suffer from complex backgrounds and deficiency of texture on the object itself. Despite that, \earl performs well in complex tasks such as human-robot handover and moving over and behind a barrier, overcoming object occlusion limitations that external camera setups typically face.

%Also, the feature detection~\cite{r2d2} and depth registration could fail on object like cup (no texture on color, highly symmetric on depth). The estimated pose rotated as the camera rotated, ending the gripper chasing the grasp pose on the cup but never closing. We only achieves $60\%$ of success rate on cup in Complex (switch) with RL method and $40\%$ with baseline method.
%
%Overall, our proposed method can achieve around $90\%$ success rate on tracking and grasping novel objects, from a table or directly from human hand. In the simulation experiments, the average success rate is $87.12\%$ for baseline and $96.27\%$ for \earl in the case of the four motions (static, line, oval, random). In the case of real robot experiments, these two numbers are $73.75\%$ and $91.25\%$, respectively. Note that we achieve a considerably small sim-to-real gap. \sj{Also write here the average success rate for random switch and Complex switch for real robot.}
% B sim  (94.9+90.76+86.54+76.28)/4=87.12
% E sim  (95.76+95+96.5+97.8)/4=96.27
% B real (85+75+65+70)/4=73.75
% E real (100+90+85+90)/4=91.25
%

\begin{table}[ht!]
    %\centering
    \vspace{1mm}
    \scalebox{0.99}{
    \hspace{-1.7mm}
    \begin{tabular}{rcc}
    \toprule
          Object Motion Type                  & \,\,\,\,\;\mtf Success\,\,\,\,\;  & \,\,\,\,\;\earl Success\,\,\,\,\;   \\ \specialrule{0.1pt}{2pt}{2pt}
         Static             & $85.00$       & \textbf{100.00}       \\ 
         Line               & $75.00$       & \textbf{90.00}       \\       
         Oval               & $65.00$       & \textbf{85.00}        \\      
         Random             & $70.00$       & \textbf{90.00}         \\ 
         Random (switch)    & $70.00$       & \textbf{90.00}        \\ 
         Complex (switch)   & $70.00$       & \textbf{80.00}        \\ 
    \bottomrule
    \end{tabular}}
    \vspace{1mm}    
    \caption{Real-world experiment results (all values in $\%$). 20 trials were performed for each test case using the UR5e real robot as the EoH system.}
    \label{tab:realtable_ur5}
\vspace{-1mm}    
\end{table}

\section{CONCLUSIONS}\label{sec:conclusion}
We presented a framework for enabling EoH systems to perform 6-DoF dynamic grasping using visual sensory inputs with active pose tracking and a curriculum-trained RL method. Our method relates sensing to actuation and applies to complex robotic tasks utilizing visual feedback for eye-in-hand control. We validated our contributions through extensive experiments in simulations and complex real-world tasks, attaining a high success rate on previously unseen objects. Our framework is generic concerning the task, but it does not consider explicit collision modeling, and the target cannot move faster than the robot. In the future, we will focus on mechanisms for recovering the target in case of tracking failures and handling more cluttered environments. Moreover, our method's performance could be enhanced by leveraging recent advancements in pose estimation~\cite{wen2023bundlesdf} and grasp detection methods.

%%%%%%%%%%%%%%%%%%%%%%%%%%%%%%%%%%%%%%%%%%%%%%%%%%%%%%%%%%%%%%%%%%%%%%%%%%%%%%%%

%%%%%%%%%%%%%%%%%%%%%%%%%%%%%%%%%%%%%%%%%%%%%%%%%%%%%%%%%%%%%%%%%%%%%%%%%%%%%%%%

%%%%%%%%%%%%%%%%%%%%%%%%%%%%%%%%%%%%%%%%%%%%%%%%%%%%%%%%%%%%%%%%%%%%%%%%%%%%%%%%

%%%%%%%%%%%%%%%%%%%%%%%%%%%%%%%%%%%%%%%%%%%%%%%%%%%%%%%%%%%%%%%%%%%%%%%%%%%%%%%%

\bibliographystyle{IEEEtran}
\bibliography{references}

\end{document}